\documentclass[10pt,twocolumn,letterpaper]{article}

\usepackage{cvpr}
\usepackage{times}
\usepackage{epsfig}
\usepackage{amsmath}

\usepackage{amssymb}
\usepackage{graphicx}
\usepackage{url}
\usepackage{cite}
\usepackage{subcaption}
\usepackage{algpseudocode}
\usepackage{algorithm}
\usepackage{mathtools}
\usepackage{amsfonts}
\usepackage{stmaryrd} % Math symbols
\usepackage{bm}
\usepackage{tikz}
\usetikzlibrary{calc,arrows,matrix,chains,positioning,decorations.pathreplacing,arrows}
\usepackage{pgfplots}
\pgfplotsset{compat = newest}
\usepackage{xspace}
\usepackage[utf8]{inputenc}
\usepackage[english]{babel}

\SetSymbolFont{stmry}{bold}{U}{stmry}{m}{n} % Avoid some warnings due to stmaryrd

\renewcommand{\vec}[1]{\boldsymbol{#1}} 

\newcommand{\pder}[2][]{\frac{\partial#1}{\partial#2}}

\newlength\figureheighttik
\newlength\figurewidthtik

\pgfplotsset{
    training_plot/.style={grid = major, grid style={dashed, gray!30}, legend entries={{Train},{Val}}, legend style={draw=none, legend columns=-1, fill=none}}
}

\usepackage{array}
\usepackage{colortbl}
\newcolumntype{L}[1]{>{\raggedright\let\newline\\\arraybackslash\hspace{0pt}}m{#1}}
\newcolumntype{C}[1]{>{\centering\let\newline\\\arraybackslash\hspace{0pt}}m{#1}}
\newcolumntype{R}[1]{>{\raggedleft\let\newline\\\arraybackslash\hspace{0pt}}m{#1}}

\newcommand{\nn}{\texttt{SegNet}\xspace}

\usepackage[pagebackref=true,breaklinks=true,letterpaper=true,colorlinks,bookmarks=false]{hyperref}

\cvprfinalcopy % *** Uncomment this line for the final submission

\def\cvprPaperID{7} % *** Enter the CVPR Paper ID here

\ifcvprfinal\fi
\begin{document}

%%%%%%%%% TITLE
\title{Deep Neural Networks for Anatomical Brain Segmentation}

\author{Alexandre de Br\'ebisson\\
Department of Mathematics, Imperial College London\\
London SW7 2AZ, UK\\
{\tt\small alexandre.de.brebisson@umontreal.ca}
\and
Giovanni Montana\\
Department of Biomedical Engineering, King's College London\\
St Thomas' Hospital, London SE1 7EH, UK\\
{\tt\small giovanni.montana@kcl.ac.uk}
}

\maketitle
\thispagestyle{empty}

%%%%%%%%% ABSTRACT
\begin{abstract}
We present a novel approach to automatically segment magnetic resonance (MR) images of the human brain into anatomical regions. Our methodology is based on a deep artificial neural network that assigns each voxel in an MR image of the brain to its corresponding anatomical region. The inputs of the network capture information at different scales around the voxel of interest: 3D and orthogonal 2D intensity patches capture a local spatial context while downscaled large 2D orthogonal patches and distances to the regional centroids enforce global spatial consistency. Contrary to commonly used segmentation methods, our technique does not require any non-linear registration of the MR images. To benchmark our model, we used the dataset provided for the MICCAI 2012 challenge on multi-atlas labelling, which consists of 35 manually segmented MR images of the brain. We obtained competitive results (mean dice coefficient 0.725, error rate 0.163) showing the potential of our approach. To our knowledge, our technique is the first to tackle the anatomical segmentation of the whole brain using deep neural networks. 
\end{abstract}

%%%%%%%%% BODY TEXT
\section{Introduction}

Quantitative research in neuroimaging often requires the anatomical segmentation of the human brain based on magnetic resonance images (MRIs). Quantitative research in neuroimaging often requires the anatomical segmentation of the human brain using magnetic resonance images (MRIs). For instance, abnormal volumes or shapes of certain anatomical regions of the brain have been found to be associated with brain disorders, including Alzheimer's disease and Parkinson~\cite{petrella2003neuroimaging, hutchinson2000structural}. The analysis of MR images is therefore essential to detect these disorders, monitor their evolution and evaluate possible treatments. The anatomical segmentation of the brain requires a segmentation protocol defining how each region should be delineated so that the resulting segmentations are comparable between brains. However, manually segmenting the brain is a time-consuming and expensive process that cannot be performed at a large scale. Its full automation would enable systematic segmentation of MRIs on the fly as soon as the image is acquired. These potential benefits have encouraged an active field of research, which is today dominated by multi-atlas based~\cite{klein2005mindboggle, heckemann2006automatic} and patch-based methods~\cite{coupe2011patch}. Machine learning methods consist in training classifiers to assign each voxel (a 3D pixel) to its anatomical region based on various input features describing it, such as its neighbourhood intensities or location. 

Recently, deep neural networks, and in particular convolutional neural networks, have proven to be the state of the art in many computer vision applications (most notably the ImageNet contest since 2012~\cite{krizhevsky2012imagenet}). Contrary to traditional shallow classifiers in which feature engineering is crucial, deep learning methods automatically learn hierarchies of relevant features directly from the raw inputs~\cite{bengio2009learning}.  Motivated by these developments, we propose a deep artificial neural network for the automated segmentation of the entire brain. This article is organised as follows. In section~\ref{background} we briefly review existing segmentation methodologies and deep neural networks. In section~\ref{methods} we describe our proposed architecture and algorithm. The application to the MICCAI 2012 dataset is presented in section~\ref{application}, and we conclude with a discussion in section~\ref{discussion}. 

\section{Background} \label{background}

Given a particular segmentation protocol, automatically segmenting a 3D MR image of the brain consists in classifying all its voxels into their corresponding protocol region. In this work, we consider the segmentation of the whole brain (cortical and sub-cortical areas) into a large number $N$ of anatomical regions, where $N$ is defined by the segmentation protocol (typically around 100). Knowledge of the segmentation protocol is implicitly given through a set of manually labelled 3D brain MRIs. An {\it atlas} consists of an MR image and its corresponding manual segmentation.

Multi-atlas based methods (such as~\cite{klein2005mindboggle, heckemann2006automatic}) are widely used methods. For a new query image to segment, such methods usually consist of the following steps: first, the $n$ most similar atlases to the query image are selected and registered to the query image;  second, the same registration transformations are applied to the labels of the $n$ atlases, and these labels are propagated to produce $n$ segmentations of the query image; finally the segmentations are combined using a fusion strategy. These methods heavily rely on a registration step, in which the atlases are non-linearly registered to the query image. A global affine or rigid registration is usually first performed and then followed by a local non-rigid registration. This latter registration step relies on the critical assumption that brains are similar enough to be accurately mapped from one to another. However, this is unlikely to be the case if the query brain is too different from the atlases in a local area (e.g. if the subject has a neurodegenerative disorder that introduces drastic structural changes). Furthermore, regions whose boundaries are clearly identifiable by a contrast in intensity but arbitrarily defined by the segmentation protocol are likely to be less accurately registered. These errors will inevitably have an effect on the final segmentation. Registrations are also computationally intensive and are usually responsible for the slowness of atlas-based methods.

In this paper we take a machine learning approach whereby, given a training set consisting of several atlases, a model is trained to classify each voxel into its corresponding anatomical region. In particular, we investigate whether recent advances in representation learning (the field of machine learning that aims to automatically extract useful representations of the data) may prove beneficial for the segmentation problem. Deep learning is the sub-field of representation learning concerned with learning multi-level or hierarchical representations of the data, where each level is based on the previous one~\cite{bengio2009learning}. Lately there has been a burst of activity around deep neural networks, and in particular convolutional neural networks, for medical imaging segmentation problems. These include approaches for the segmentation of the lungs~\cite{middleton2004segmentation}, cells of C. elegans embryos~\cite{ning2005toward}, biological neuron membrane~\cite{ciresan2012deep}, tibial cartilage~\cite{prasoon2013deep}, bone tissue~\cite{cernazanu2013segmentation} and cell mitosis~\cite{cirecsan2013mitosis}, amongst others. All these applications mostly use 2D convolutional networks which take intensity patches as inputs; occasionally spatial consistency is enforced at a second stage through post-processing computations such as probabilistic graphical models. Despite this increasing interest in medical imaging, deep neural networks have not yet been considered for the problem of the whole brain segmentation into anatomical regions. Initial work has been carried out for the segmentation of a single (central) 2D slice of the brain using local 2D patches as input \cite{lee2011towards}. By comparison, our approach tackles the segmentation of the whole 3D brain and introduces multi-scale input features to enforce the spatial consistency of the segmentation.

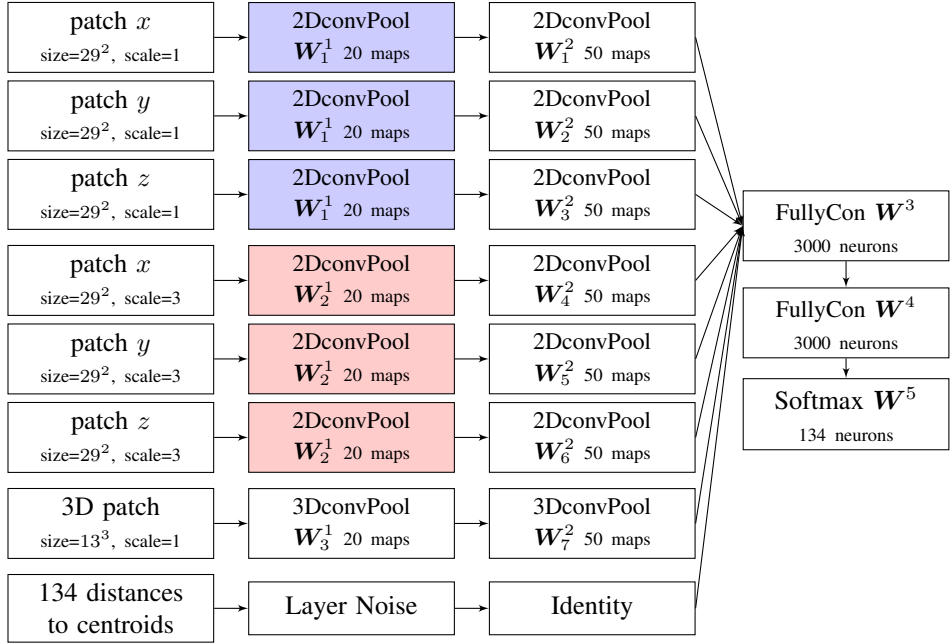
\begin{figure*}[!ht]
	\centering
	\def\a{0.1cm}
\def\b{0.2cm}
\def\c{3.2cm}
\def\d{2cm}

\tikzstyle{inputs}=[draw, minimum size=2em, text width=2.5cm, align=center]
\tikzstyle{layer1}=[inputs, fill=blue!20]
\tikzstyle{layer2}=[inputs]
\tikzstyle{layer3}=[inputs]
\tikzstyle{layer4}=[inputs]
\tikzstyle{init} = [pin edge={to-,thin,black}]

\begin{tikzpicture}[node distance=\c,auto,>=latex']

% 3 orthogonal scale1
	\node [inputs] (I1) {patch $x$ \\ \scriptsize size=$29^2$, scale=1};
	\node [inputs, below=\a of I1] (I2) {patch $y$ \\ \scriptsize size=$29^2$, scale=1};
	\node [inputs, below=\a of I2] (I3) {patch $z$ \\ \scriptsize size=$29^2$, scale=1};
    
    \node [layer1, right of=I1] (conv1) {\small 2DconvPool \\ $\vec{W}_1^1$ \scriptsize 20 maps};
    \node [layer1, right of=I2] (conv2) {\small 2DconvPool \\ $\vec{W}_1^1$ \scriptsize 20 maps};
    \node [layer1, right of=I3] (conv3) {\small 2DconvPool \\ $\vec{W}_1^1$ \scriptsize 20 maps};

    \path[->] (I1) edge node {} (conv1);
    \path[->] (I2) edge node {} (conv2);
    \path[->] (I3) edge node {} (conv3);
        
    \node [layer2, right of=conv1] (conv11) {\small 2DconvPool \\ $\vec{W}_1^2$ \scriptsize 50 maps};
    \node [layer2, right of=conv2] (conv22) {\small 2DconvPool \\ $\vec{W}_2^2$ \scriptsize 50 maps};
    \node [layer2, right of=conv3] (conv33) {\small 2DconvPool \\ $\vec{W}_3^2$ \scriptsize 50 maps};
  
    \path[->] (conv1) edge node {} (conv11);
    \path[->] (conv2) edge node {} (conv22);
    \path[->] (conv3) edge node {} (conv33);

%three orthogonal scale=3
	\node [inputs, below=\b of I3] (I4) {patch $x$ \\ \scriptsize size=$29^2$, scale=3};
	\node [inputs, below=\a of I4] (I5) {patch $y$ \\ \scriptsize size=$29^2$, scale=3};
	\node [inputs, below=\a of I5] (I6) {patch $z$ \\ \scriptsize size=$29^2$, scale=3};
    
    \node [layer1, fill=red!20, right of=I4] (conv4) {\small 2DconvPool \\ $\vec{W}_2^1$ \scriptsize 20 maps};
    \node [layer1, fill=red!20, right of=I5] (conv5) {\small 2DconvPool \\ $\vec{W}_2^1$ \scriptsize 20 maps};
    \node [layer1, fill=red!20, right of=I6] (conv6) {\small 2DconvPool \\ $\vec{W}_2^1$ \scriptsize 20 maps};

    \path[->] (I4) edge node {} (conv4);
    \path[->] (I5) edge node {} (conv5);
    \path[->] (I6) edge node {} (conv6);
        
    \node [layer2, right of=conv4] (conv44) {\small 2DconvPool \\ $\vec{W}_4^2$ \scriptsize 50 maps};
    \node [layer2, right of=conv5] (conv55) {\small 2DconvPool \\ $\vec{W}_5^2$ \scriptsize 50 maps};
    \node [layer2, right of=conv6] (conv66) {\small 2DconvPool \\ $\vec{W}_6^2$ \scriptsize 50 maps};
  
    \path[->] (conv4) edge node {} (conv44);
    \path[->] (conv5) edge node {} (conv55);
    \path[->] (conv6) edge node {} (conv66);

% 3d patch
	\node [inputs, below=\b of I6] (I7) {3D patch\\ \scriptsize size=$13^3$, scale=1};
	
    \node [layer2, right of=I7] (conv7) {\small 3DconvPool \\ $\vec{W}_3^1$ \scriptsize 20 maps};	
    
    \path[->] (I7) edge node {} (conv7);
    
    \node [layer2, right of=conv7] (conv77) {\small 3DconvPool \\ $\vec{W}_7^2$ \scriptsize 50 maps};
	
	\path[->] (conv7) edge node {} (conv77);

% Centroids
	\node [inputs, below=\b of I7] (I8) {134 distances to centroids};
	
	\node [inputs, right of=I8] (conv8) {Layer Noise};
	
	\path[->] (I8) edge node {} (conv8);
	
	\node [inputs, right of=conv8] (conv88) {Identity};
	
	\path[->] (conv8) edge node {} (conv88);

 	\coordinate (p1) at ($(conv11)!0.33!(conv88)$);
 	\coordinate (p2) at ($(conv11)!0.5!(conv88)$);
 	\coordinate (p3) at ($(conv11)!0.66!(conv88)$);
    \node [layer3, right=\d of p1] (l3) {\small FullyCon $\vec{W}^3$ \\ \scriptsize 3000 neurons};
    
    \path[->] (conv11.east) edge node {} (l3.west);
    \path[->] (conv22.east) edge node {} (l3.west);
    \path[->] (conv33.east) edge node {} (l3.west);
    \path[->] (conv44.east) edge node {} (l3.west);
    \path[->] (conv55.east) edge node {} (l3.west);
    \path[->] (conv66.east) edge node {} (l3.west);
    \path[->] (conv77.east) edge node {} (l3.west);
    \path[->] (conv88.east) edge node {} (l3.west);
 
    \node [layer3, right=\d of p2] (l4) {\small FullyCon $\vec{W}^4$ \\ \scriptsize 3000 neurons}; 
    
	\path[->] (l3) edge node {} (l4);    
    
    \node [layer3, right=\d of p3] (l5) {Softmax $\vec{W}^5$ \\ \scriptsize 134 neurons};
    
    \path[->] (l4) edge node {} (l5);
      
\end{tikzpicture}
	\caption{Architecture of \nn optimised for the MICCAI 2012 dataset. There are 8 pathways, one for each input feature. The lower layers (on the left-hand side) learn specific representations of their input features, which are then merged into a joint representation. Each {\tt 2DconvPool} block represents a convolutional layer followed by a pooling layer. The colour indicates that the layer blocks share the same parameters. The distances to centroids are estimated as explained in section~\ref{estimation_centroids}. The parameter values (patch sizes, scales, number of neurons) shown here are those selected for the MICCAI application, in which the convolutional layers have $5 \times 5$ kernels and $2 \times 2$ max-pooling windows. The noise layer is only activated during training, and the model has a total of $30,565,555$ parameters.}
	\label{net:ultimate}
\end{figure*}

\section{Architecture of the network} \label{methods}

In this section, we describe the inputs and the architecture of the proposed network.

\subsection{Input Features}

We aim at designing an algorithm that classifies each voxel into its corresponding anatomical region. Each voxel must therefore be described by an input vector, which is the input of our neural network. The choice of input is particularly important as it should capture enough information for the task while being as parsimonious as possible, mostly for computational reasons and to avoid overfitting.  Two types of inputs were incorporated in this work in order to ensure both local precision and global spatial consistency.

\subsubsection{Features to ensure local precision.}

For each voxel, the local precision of the segmentation is ensured by the two following features.  First, a 3D patch of size $a \times a \times a$ centred on the voxel is used to capture local information at a high level of detail. Second, three 2D orthogonal patches of size $b \times b$ (each extracted from the sagittal, coronal and transverse planes respectively), also centred on the voxel, are added with the purpose of capturing a slightly broader but still local context around the voxel of interest. The use of these orthogonal patches can be seen as a trade-off between a single 2D patch and a 3D patch: they capture 3D information but require a significantly smaller amount of memory for storage than a dense 3D patch, allowing bigger patch sizes to be used.

\subsubsection{Features to ensure global spatial consistency.}

\begin{figure}[!ht]
	\centering
	\includegraphics[width=6.5cm]{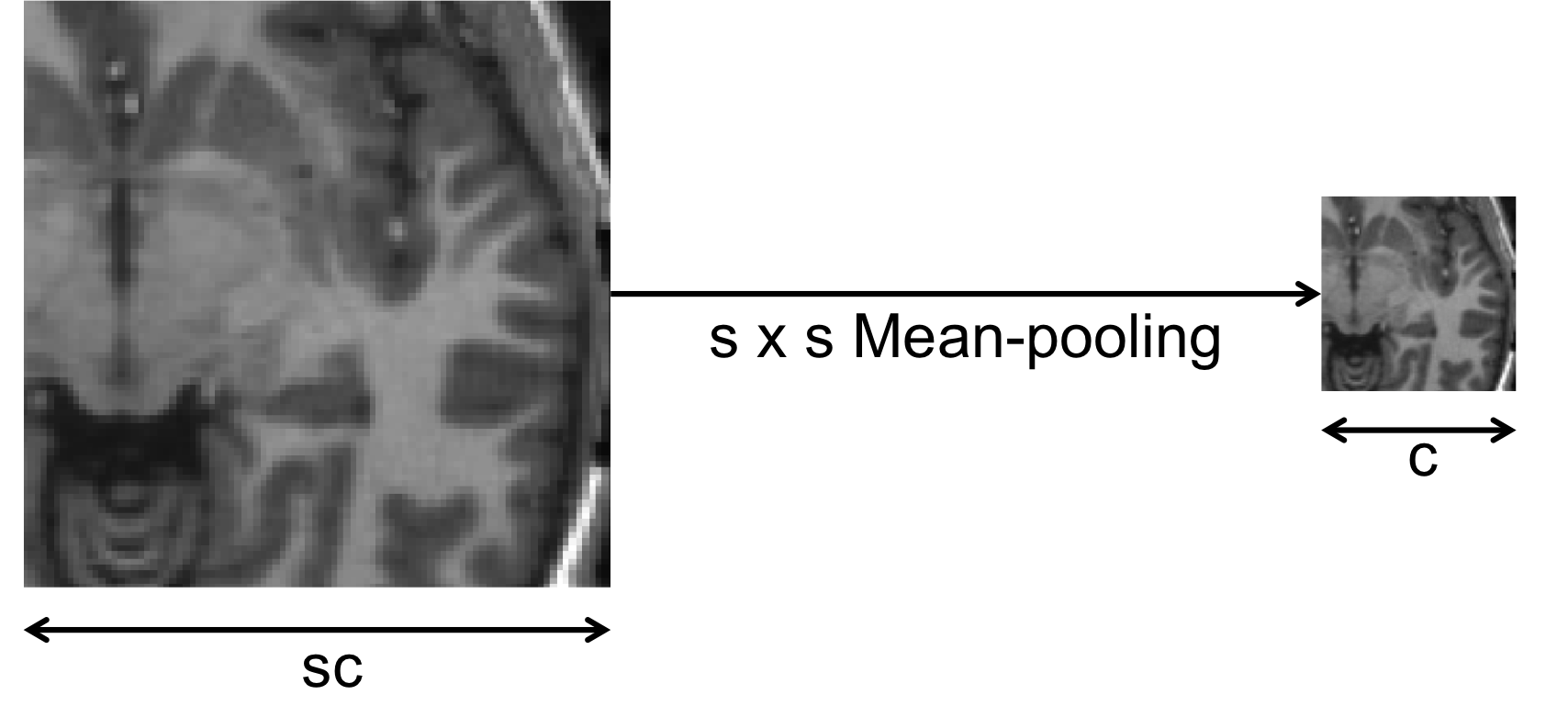}
	\caption{A downscaled patch spans the same region of the MRI as the original patch but with a lower resolution.}
	\label{fig:downscale}
\end{figure}

A second set of inputs was designed to preserve global spatial consistency. Unlike unstructured segmentation tasks, in which different regions can be arbitrarily positioned in an image, anatomical regions consistently preserve the same relative positions in all the subjects. Including global information is therefore likely to yield additional improvements. An obvious strategy would be to simply increase the size of the 2D and/or 3D patches introduced earlier so as to span larger portions of the image and cover more distant anatomy. However this would generate very high-dimensional inputs requiring large memory for storage and would add computational complexity. Instead, we extract large 2D orthogonal patches that we \emph{downscale} by a factor $s$. As illustrated in figure~\ref{fig:downscale}, the downscale operation simply reduces the resolution of the patch by averaging voxel intensities within small square windows of size $s \times s$. More precisely, if $sc \times sc$ is the size of the original full-resolution patches, then the downscaled patches have sizes $c \times c$. In neural network terminology, this operation is equivalent to a $s \times s$ mean-pooling with stride $s$. As a result, the downscaled patch still captures as large portions of the MRI as the original patch but with lower resolution.

In addition to the voxel intensities, the coordinates of each voxel in 3D space are also expected to be very informative for anatomical segmentation purposes. However the use of absolute coordinates is predicated on the individual brain scans to be represented in a common reference space, which in turn requires performing an initial -- and generally very time consuming -- registration of all the images.  As an alternative, we explore the use of relative distances from each voxel to each one of the $N$ centroids as additional inputs of the network. The centroid $c_l=(x_l,y_l,z_l)$ of region $l$ of image $I$ is defined as the center of mass of all the uniformly weighted voxels of that region:
\[
c_l = \frac{\sum_{v \in I^{-1}(l)} v }{|I^{-1}(l)|},
\]
where $I^{-1}(l)$ represents the set of all voxels belonging to region $l$. The distance $d$ -- simply taken to be Euclidian -- of a voxel to all the region centroids gives an indication of the position of the voxel in the image and thus the region it belongs to. Unlike absolute coordinates, these distances are invariant to rotations and translations. The distances to centroids are also invariant to brain scaling upon scaling the centroids and the image coordinates by the average distance between two centroids, $D$, defined as:
\[
D = \frac{N \times (N+1)}{2} \sum_{i=1}^{N} \sum_{j=i}^{N} d(c_i, c_j).
\]
This enforces the average distance between two centroids to be the same for all the brains. In practice, only a few correct distances from a voxel to the centroids would be sufficient to localise precisely the current voxel, but by adding all distances we increase the robustness of the resulting segmentation to noise, which is confirmed by our experimental results. Although these distances can be computed exactly using the training data set, for which all the centroids are known, they are unknown on new brains before any segmentation has  been performed. To deal with this issue, we propose a two-stage algorithm, in which the first stage provides a segmentation of the brain without using these distances, and then a second, refinement stage is added to further improve the segmentation; see section~\ref{estimation_centroids} for further details.

\subsection{Deep Neural Network}

Our proposed network architecture, called \nn, is represented in Figure~\ref{net:ultimate}. It is a feed-forward network formed by stacking $K$ layers of artificial neurons. Each layer models a new representation of the data, in which neurons act as feature detectors. Recursively, deeper neurons learn to detect new features formed by those detected by the previous layer. The result is a hierarchy of higher and higher level feature detectors. This is natural for images as they can be decomposed into edges, motifs, parts of regions and regions themselves. We let $\nn^l$ denote the function mapping the inputs of layer $l$ to its output. Our architecture is then expressing a function $\nn_{\vec{\theta}}$ (or more simply $\nn$) defined as
\[ \nn_{\vec{\theta}} = \nn^K \circ \cdots \circ \nn^l \circ \cdots \circ \nn^1,\]
where $\vec{\theta}$ represents the parameters of the network, i.e. the weights and the biases.

Our network architecture has 8 types of input features and 8 corresponding pathways that are merged later in the network. The lowest layers are specific to each type of input features and aim to learn specialised representations. Apart from the distances to centroids, these representations are learnt by 2D and 3D convolutional~\cite{lecun1998gradient} and pooling layers~\cite{ranzato2007unsupervised}, denoted {\tt 2DconvPool} and {\tt 3DconvPool}, respectively. At a higher layer of this architecture, the individual representations are merged into a common representation across all the inputs. Further layers learn even higher level representations, which capture complex correlations across the different input features. These representations are learnt with fully connected layers, denoted {\tt FullyCon}. 

\subsubsection{Convolutional Layers.}

2D and 3D patches are processed with convolutional layers~\cite{lecun1998gradient}, which aim to detect local features at different positions in an image. The neurons of a convolutional layer compute their outputs based only on a subset of the inputs, called the receptive field of the neuron. More precisely, a neuron in a given convolutional layer depends only on a spatial contiguous set of the layer inputs, in our case $t \times t$ windows of voxel intensities. Therefore each neuron learns a particular local feature specific to its receptive field. This local connectivity considerably reduces the number of parameters and thus the potential overfitting of the layer. In addition to local connectivity, a convolutional layer also imposes groups of neurons, called feature maps, to share exactly the same weight values. More precisely, a convolutional layer can be decomposed into several feature maps, whose neurons share the same weights and only differ by their receptive field. This means that neurons of a same feature map detect the same feature but from different receptive fields of the image. The local connectivity and weight sharing constraints can simply be modelled by a  sum of convolution operations. The outputs of neurons in feature map $k$ of layer $l$ is then given by
\[\vec{h}^{l}_k = \varphi \left( \sum_{m} \vec{W}^{l}_{m,k} * \vec{h}^{l-1}_{m} + b^l_k \right), \]
where $*$ is the convolution operation and $\vec{W}^{l}_{m,k}$ is the weight matrix for the feature map $k$ of layer $l$ and feature map $m$ of layer $l-1$ (the kernel of the convolution). $\vec{W}^{l}_{m,k}$ is a 2D or a 3D matrix respectively for 2D or 3D images. Its size is the size of the receptive fields of each neuron in the feature map. Here $\vec{h}^{l-1}_{m}$ is the feature map $m$ of layer $l-1$, and $b^l_k$ is the scalar bias of the feature map $k$ of layer $l$.

\subsubsection{Max-Pooling Layers.}

The convolutional layers of our architecture are followed by max-pooling layers~\cite{ranzato2007unsupervised}, which reduce the size of the feature maps by merging groups of neurons. More precisely, for each datapoint, a max-pooling layer shifts a square window ($p \times p$ in our case) over the feature map and select the most responsive neuron over each position of the window, the other neurons being discarded. The output of the most responsive neuron indicates if the feature map has detected its corresponding feature in the receptive field of the pooling window, the precise receptive field of the feature being lost. As the local information is particularly important in our problem, we only considered small windows. The benefits of pooling layers is that they significantly reduce the number of parameters, making the training simpler and reducing overfitting.

\subsubsection{Fully Connected Layers.} 

The latest layers of our architecture are fully connected layers, denoted {\tt FullyCon}. The output vector $\vec{h}^l$ of a fully connected layer $l$ is given by
\[ \vec{h}^l = {\nn^l}(\vec{h}^{l-1}) = \varphi(\vec{W}^l \vec{h}^{l-1} + \vec{b}^l), \]
where $\vec{h}^{l-1}$ is the input of the layer $l$ and where $\varphi$ is the activation function of the layer.

\subsubsection{Activation Functions.}

Apart from the top layer, we used the same activation function for all the neurons of our network, the rectifier function defined by
$$
\varphi: x \mapsto \max(0,x).
$$
A neuron with the rectifier function is called a REctified Linear Unit (RELU). Contrary to the more traditional sigmoid or $\tanh$ functions, it is less prone to the vanishing gradient problem, which has prevented the training of deep networks for several decades~\cite{glorot2011deep}.

The top layer of our network architecture uses a softmax activation function. If $z_r$ is the weighted input of output neuron $r$, then the output of neuron $j$ is given by
\[
output_j = \frac{e^{\mathbf{z}_j}}{\sum_{r=1}^{N} e^{\mathbf{z}_r}}.
\]
The softmax function maps the weighted inputs into $[0,1]$ and the outputs can then be interpreted as probabilities. In practice, to label a voxel, we select the output with the highest probability.

\subsubsection*{Weight Sharing}

The reason for sharing the weights $\vec{W}_1^1$ (corresponding to the three orthogonal 2D patches) and $\vec{W}_2^1$ (corresponding to the three orthogonal 2D downscaled patches) between layer blocks in the lowest layer is that we believe that the lowest level features that the network learns about the patches should be the same whatever the orientation of the patch. This also divides the number of parameters in the first layer by two, which reduces the risk of overfitting. Experimentally, we have found that this constraint slightly improves the performance.

\subsection{Training Algorithm}

Let us denote the training dataset by $\left\lbrace (\vec{x}^{(i)}, \vec{y}^{(i)})~|~i \in \llbracket 1, n \rrbracket \right\rbrace$, where for each $i$, $\vec{y}^{(i)}$ is the known desired output of input $\vec{x}^{(i)}$ ($\vec{y}^{(i)}$ is a vector of zeros with a single one at the position of the classification label). The performance of the network is evaluated using negative log-likelihood error function, which is defined as follows
\[ E_{CE}: \vec{\theta} \mapsto - \frac{1}{n} \sum_{i=1}^n \log \left( \nn_{\vec{\theta}}(\vec{x}^{(i)}) \cdot \vec{y}^{(i)} \right), \]
where $\cdot$ is the dot product in the output space defined as $a \cdot b = \sum_{j=1}^N a_j b_j$ and $\vec{\theta}$ represents all the parameters of the network. Training is cast into the minimisation of $E_{CE}$, which was carried out by the stochastic gradient descent (SGD) algorithm, a variant of the gradient descent algorithm commonly used to train large networks on large datasets~\cite{bousquet2008tradeoffs}. At each update of the weights in the SGD algorithm, instead of considering all the training datapoints to compute the gradient of the error function $E_{CE}$, only one datapoint or a small batch of training datapoints is used. We also added a momentum term~\cite{rumelhart1988learning,sutskever2013importance}, which is particularly beneficial along long narrow valleys of the error function as it averages the directions of the gradient. If $\Delta w_{ij}^l(t)$ denotes the update of weight $w_{ij}^l$ at iteration $t$, then the momentum update rule is given by
\[ \Delta w_{ij}^l(t) = -\alpha \pder[E]{w_{ij}^l} + m \Delta w_{ij}^l(t-1), \]
where the scalars $\alpha$ and $m$ are the learning rate and the momentum, respectively.

\label{estimation_centroids}
\subsection{Estimation of the Centroids}

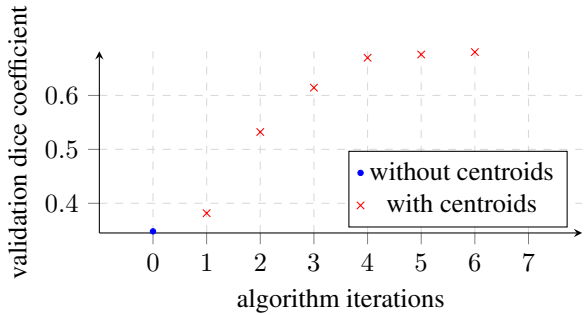
\begin{figure}[!ht]
	\centering
	\begin{tikzpicture}
    \begin{axis}[
    	axis lines=left,
    	legend pos=south east,
        grid = major,
        width=8cm,
        height=4cm,
        grid style={dashed, gray!30},
        xmin= -1,     % start the diagram at this x-coordinate
        xmax= 8,    % end   the diagram at this x-coordinate
        ymin= 0.345,     % start the diagram at this y-coordinate
        ymax= 0.682,   % end   the diagram at this y-coordinate
        %axis background/.style={fill=white},
        xlabel=algorithm iterations,
        ylabel=validation dice coefficient,
        tick align=outside,
        enlargelimits=false,
        xtick = {0,...,7}]
        
	\addplot[only marks, color=blue,mark=*, mark size=1] coordinates {
		(0,0.348)
	};
	\addplot[only marks, color=red,mark=x, mark size=2] coordinates {
		(1,0.38166)
		(2,0.53219)
		(3,0.61442)
		(4,0.66985)
		(5,0.67589)
		(6,0.68034)
	};
      \addlegendentry{without centroids}
      \addlegendentry{with centroids}
	\end{axis}
\end{tikzpicture}
	\caption{Example of the convergence of the algorithm for a toy fully-connected network, trained on the MICCAI dataset~\ref{application}, with poor initial accuracy. In this particular case, the approximated distances to centroids, which are updated at each iteration, significantly improve the segmentation.}
	\label{fig:centroids_convergence}
\end{figure}

Training \nn using the distances to centroids is possible as the true segmentations are available. However, when we consider the segmentation of a new brain, we can not compute directly the centroids. Therefore we propose the following iterative procedure that uses two neural networks in sequence. A sub-network, which corresponds to \nn without the centroids pathway, is first trained. Given a new MR image, this sub-network is used to produce an initial segmentation of the image, which enables to compute approximated centroids of each region and then distances between each voxel and these approximated centroids. The full \nn network, now taking the centroids pathways as additional input, is then used to obtain a refined segmentation. This refined segmentation is then used to compute a better approximation of the centroids. These two last steps can be repeated multiple times until convergence, i.e. no changes in the segmentation output. In practice, we observed that, when the initial model is already relatively accurate, the algorithm converges really fast in only a few iterations. When the initial model has poor segmentation accuracy, the algorithm is slightly longer to converge, as illustrated in figure~\ref{fig:centroids_convergence}. When the initial model is really too bad, the algorithm does not improve the initial accuracy.

Even though the centroids are only approximated, distances carry sufficient information to identify the region in which each voxel lies. To enforce robustness to noisy approximations of the centroids, during the training stage of the neural network, we artificially corrupt the distances to centroids by adding gaussian noise for each training datapoint. This makes sure that the network relies less on individual distances but more on the statistical properties of the group of $N$ distances.

\section{An application to the MICCAI dataset} \label{application}

We tested our approach on the dataset of the MICCAI 2012 challenge on multi-atlas labelling. At the beginning of this competition, the organisers released $15$ atlases and $20$ MRIs (around $200^3$ pixels) without segmentations. Each team was required to develop a segmentation algorithm using the $15$ atlases and submit their segmentations for the $20$ MRIs. The quality of the segmentation was assessed by computing the mean dice coefficient over the anatomical regions. The $35$ images in this dataset are T1-weighted structural MRIs obtained from the OASIS project~\cite{marcus2007open} . These images have been manually aligned using translation and rotation, and were segmented by NeuroMorphometric into $134$ anatomical regions. The non-cortical regions follow  the NeuroMorphometric protocol\footnote{\url{http://neuromorphometrics.org:8080/Seg/}}, while the cortical regions follow the BrainCOLOR protocol\footnote{\url{http://www.braincolor.org/protocols/cortical\_protocol.php}}. The winning team of the MICCAI challenge obtained an overall mean dice coefficient of $0.765$ and the median coefficient over all the teams was $0.7251$.

\begin{figure}[t]
	\centering
	\begin{subfigure}[t]{0.99\columnwidth}
	\centering
	\includegraphics[width=5.8cm]{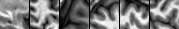}
	\caption{$29 \times 29$ patches ($b=29$).}
	\end{subfigure}
	
	\begin{subfigure}[t]{0.99\columnwidth}
	\centering
	\includegraphics[width=5.8cm]{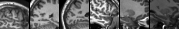}
	\caption{$87 \times 87$ original patches downscaled into $29 \times 29$ patches ($c=29$, $s=3$).}
	\end{subfigure}

	\caption{Example of random 2D patches from the MICCAI dataset.}
	\label{fig:patch_2d}
\end{figure}

\begin{figure}[t]
	\centering
	\begin{subfigure}[t]{0.42\columnwidth}
	\centering
	\includegraphics[width=3.5cm]{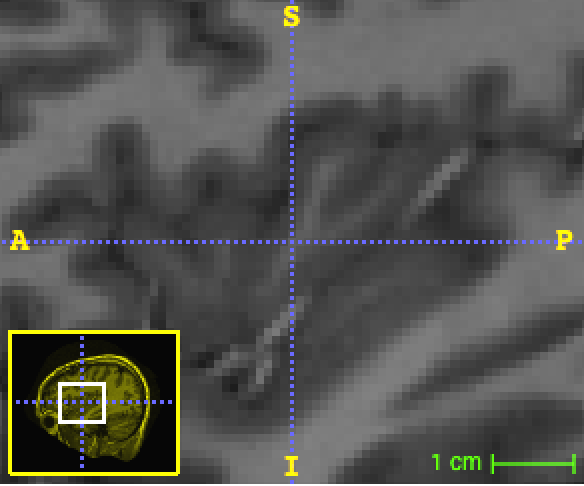}
	\end{subfigure}
	\begin{subfigure}[t]{0.42\columnwidth}
	\centering
	\includegraphics[width=3.5cm]{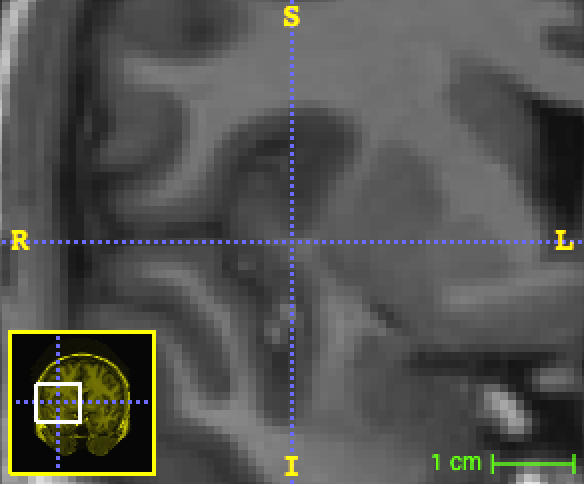}
	\end{subfigure}
	\begin{subfigure}[t]{0.42\columnwidth}
	\centering
	\includegraphics[width=3.5cm]{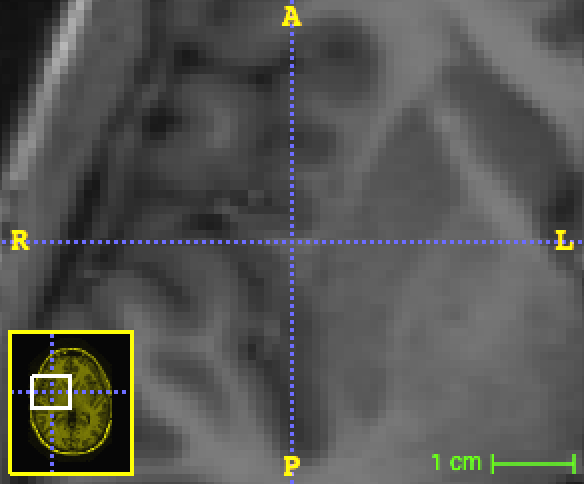}
	\end{subfigure}
	
	\caption{Example of three orthogonal patches centred on the voxel of interest. Screenshots captured with ITK-SNAP.}
	\label{fig:patch_2d}
\end{figure}

We set out to study the performance of \nn on this dataset despite the fact that deep learning techniques generally require much larger training datasets. All computations had to be run in-memory using a single NVIDIA Tesla K40 GPU with 12GB memory.  Therefore we faced a trade-off between the number of datapoints and the number of dimensions of the dataset. On the basis of initial tests, we decided to extract randomly and uniformly across the brain a sample of approximately $20$k voxels from each one of the $15$ atlases, for a total of $300$k voxels for training purposes, which amounts to only approximatively $1.5\%$ of all the available voxels in the dataset. For each voxel, we extracted a $7377$-dimensional input vector consisting of a 3D patch of $13^3$ voxel intensities ($a=13$), three 2D orthogonal patches of $29^2$ voxel intensities ($b=29$), three 2D downscaled patches of size $29^2$ containing averaged voxel intensities (the original patch width is $87$ and the scale is $3$, i.e. $c=29$, $s=3$) and $134$ distances to centroids. Figure~\ref{fig:patch_2d} shows a sample of 2D patches. A validation dataset consisting of $40$k data points was also extracted from the the 15 atlases. 

Early stopping was applied to the error rate on the validation dataset with a patience of 10 epochs. Although we were primarily interested in the dice coefficient, we applied early stopping to the error rate, which is more stable than the dice coefficient as a single misclassification (especially if it occurs for a small region) impacts more the latter than the former. The final architecture has 7 layers ($K=7$). The sizes of the convolution kernels were set to $5 \times 5$ ($t=5$) and those of the pooling windows to $2 \times 2$ ($p=2$). The other parameters are reported in Figure~\ref{net:ultimate}. The resulting architecture has a total of $30,565,555$ parameters. We used a batch size of $200$ data points. The momentum and learning rate were both tuned and set respectively to $m=0.5$ and $\alpha=0.05$. The code\footnote{Available at \url{https://github.com/adbrebs/brain_segmentation}} is based on Theano~\cite{bergstra+al:2010-scipy}, a python library that compiles symbolical expressions into C/CUDA code that can run on GPUs.

\begin{table}[!ht]
\centering
\arrayrulecolor{gray!30}
\begin{tabular}{L{3.6cm} !{\color{black}\vrule}!{\color{black}\vrule} C{1.5cm} | C{1.5cm} }
\textbf{Model} & \textbf{Number of inputs} & \textbf{Val error} \\ [0.5ex] 
\arrayrulecolor{black}\hline\hline\arrayrulecolor{gray!30}
\multicolumn{3}{c}{Local precision features} \\\hline
2D patch & $29^2$ & 0.607 \\\hline
3D patch & $13^3$ & 0.303 \\\hline
3 2D patches & $3 \times 29^2$ & 0.175 \\
\arrayrulecolor{gray}\hline
\hline\arrayrulecolor{gray!30}
\multicolumn{3}{c}{Spatial consistency features} \\\hline
3 2D downscaled patches & $3 \times 29^2$ & 0.207\\\hline
coordinates $(x,y,z)$ & $3$ & 0.410 \\\hline
distances to centroids  & $134$ & 0.305 \\
\arrayrulecolor{gray}\hline\hline\arrayrulecolor{gray!30}
\multicolumn{3}{c}{Combining features} \\\hline
3 2D patches +\\ 3 2D downscaled patches & $6 \times 29^2$ & 0.124 \\\hline
\nn & $6 \times 29^2 + 13^3 + 134$ & 0.105 \\
\end{tabular}
\caption{Validation errors of models optimized for individual and combined input features. \nn is the network represented in Fig.~\ref{net:ultimate}.}
\label{tab:res}
\end{table}

Table~\ref{tab:res} shows the validation error rate for various model architectures that only include one input at a time, and whose parameters have been optimised on the training data. Here it can be observed that the combined use of three 2D orthogonal patches dramatically improves the segmentation performance compared to 2D or 3D patches. We can also notice that the distances to centroids, in addition to their invariance qualities, significantly outperform the coordinates $(x,y,z)$. We also observed that, for already manually segmented brains, using estimated centroids yield equivalent results as using the true centroids. Our best model, \nn, was selected on the basis of the validation error ($0.105$). Evaluated on the 20 testing MRIs of the MICCAI challenge, it has a mean dice coefficient of $0.725$ and an error rate of $0.163$. The 20 testing MRIs were never used during the training nor the selection of architectures. 

\begin{figure}[!ht]
	\centering
	\begin{subfigure}[t]{0.32\columnwidth}
	\centering
	\includegraphics[width=2cm]{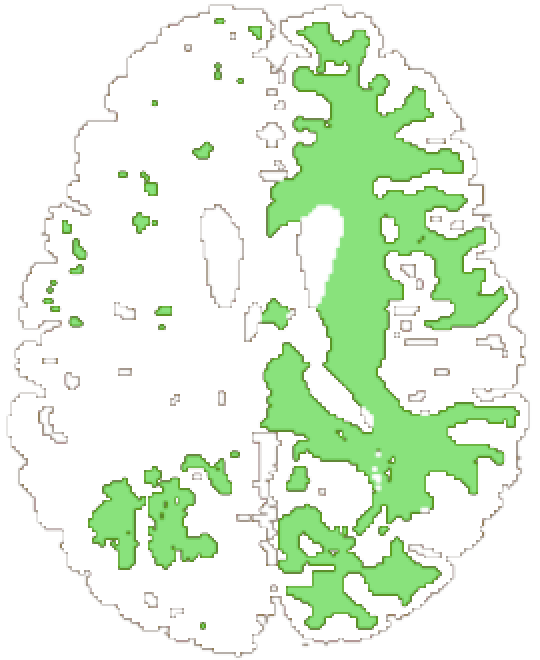}
	\caption{}
	\label{subfig:fig1}
	\end{subfigure}
	\begin{subfigure}[t]{0.32\columnwidth}
	\centering
	\includegraphics[width=2cm]{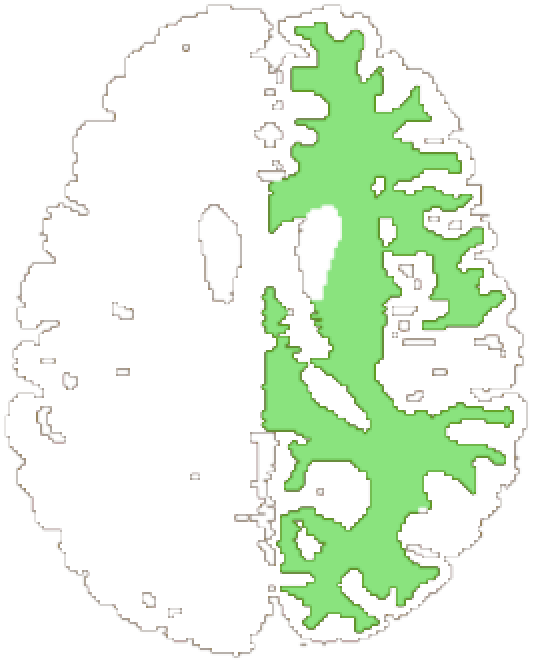}
	\caption{}
	\label{subfig:fig2}
	\end{subfigure}
	\begin{subfigure}[t]{0.32\columnwidth}
	\centering
	\includegraphics[width=2cm]{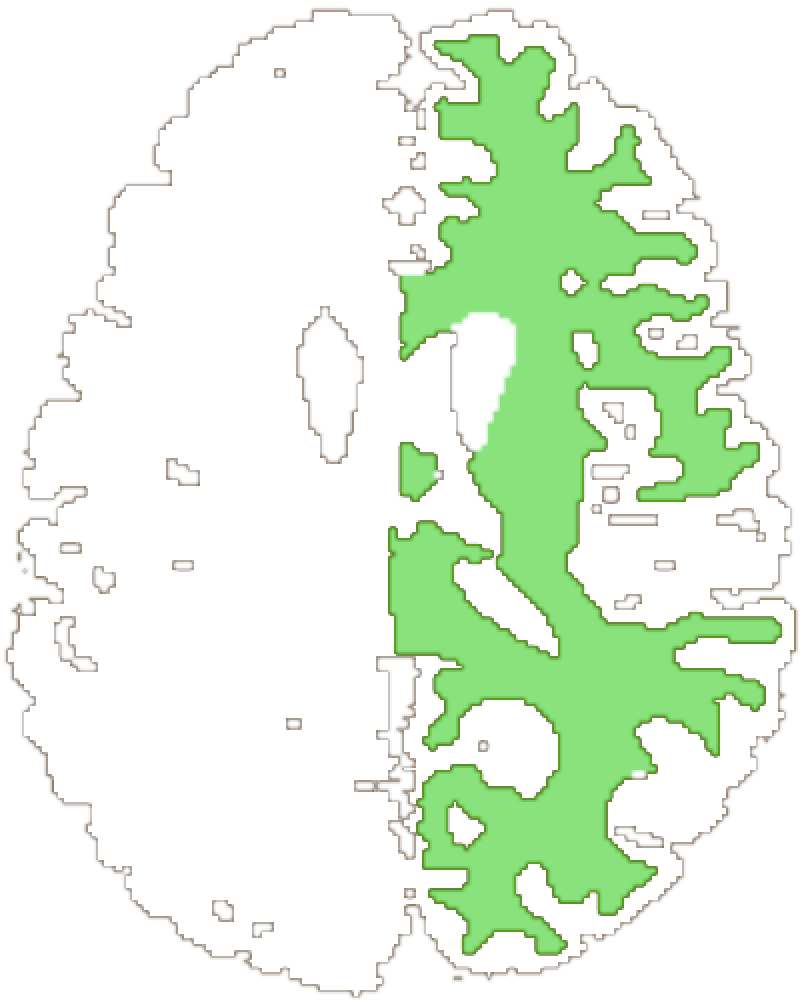}
	\caption{}
	\label{subfig:fig3}
	\end{subfigure}
	\caption{Comparison of the automatic and manual segmentations of the cerebral white matter region (in green) of the right hemisphere of an example MRI (ID: 1004). (\subref{subfig:fig1}) Segmentation returned by the network using only the three orthogonal 2D patches as inputs, (\subref{subfig:fig2}) Manual segmentation, (\subref{subfig:fig3}) Segmentation returned by \nn. Contrary to (\subref{subfig:fig3}), (\subref{subfig:fig1}) wrongly classifies some parts of the left hemisphere.
}
	\label{fig:spatial_consistency_comparison}
\end{figure}

\begin{figure}[!ht]
	\centering
	\begin{subfigure}{\columnwidth}
		\centering
		\includegraphics[width=0.7\textwidth]{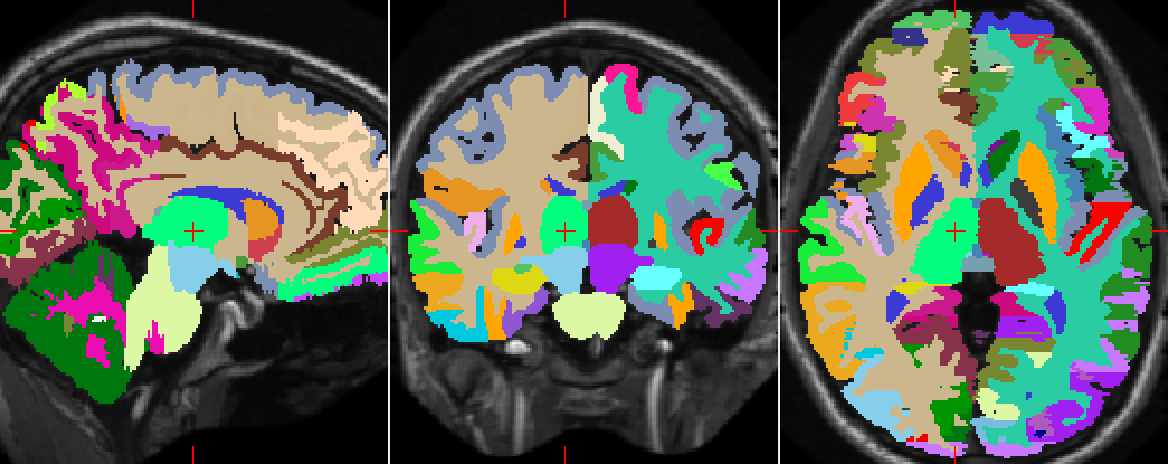}
		\caption{Manual segmentation.}
	\end{subfigure}	

	\begin{subfigure}{\columnwidth}
		\centering
		\includegraphics[width=0.7\textwidth]{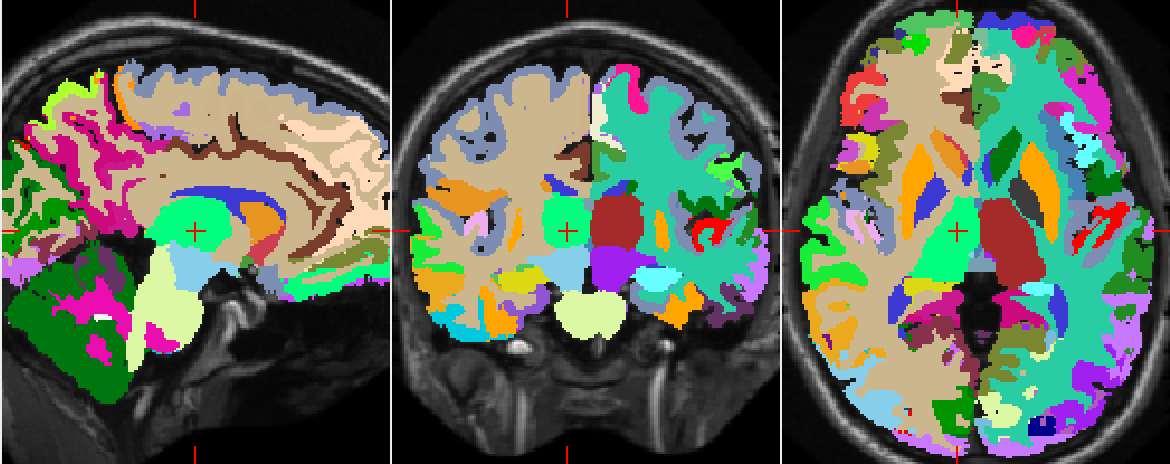}
		\caption{Predicted segmentation.}
	\end{subfigure}
	
	\begin{subfigure}{\columnwidth}
		\centering
		\includegraphics[width=0.7\textwidth]{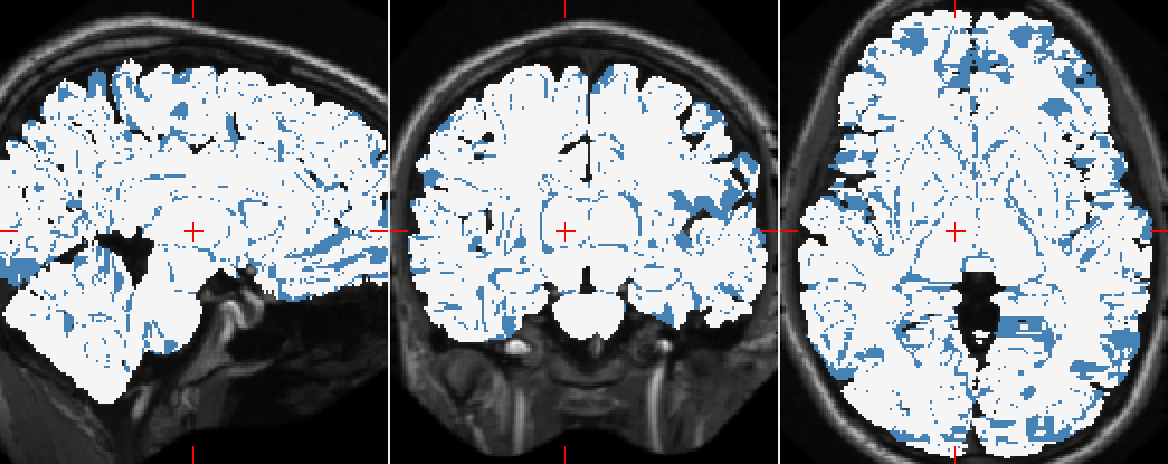}
		\caption{Difference, white voxels are identical while blue are different.}
		\label{brain:ultimate_diff}
	\end{subfigure}	
	\caption{Comparison of the manual brain segmentation of a subject (ID: 1004) to that predicted by \nn. The mean dice coefficient is 0.74.}
	\label{brain:ultimate}
\end{figure}

Figure~\ref{fig:spatial_consistency_comparison} illustrates how well the downscaled patches and the distances to centroids enforce the global spatial consistency of the segmentations. Figure~\ref{brain:ultimate} shows the manual and automatic segmentations of a particular MRI. We notice that the misclassified voxels tend to lie on the boundaries of the regions, as expected.

\section{Conclusion} \label{discussion}

We designed a deep neural network architecture to automatically segment brain MRIs. We benchmarked it against the multi-atlas methods of the MICCAI challenge of 2012 and obtained competitive results (mean dice $0.725$). Contrary to multi-atlas based methods, ours do not rely on any non-linear registrations of the MRIs. Therefore, although this has not been verified in the context of this work, we also expect our method to generalise better when the query image exhibits an abnormal region volume or shape that is not represented in the training atlases. 

We proposed two types of input features and a corresponding architecture to precisely delineate the boundaries of the regions while ensuring global spatial consistency. Due to current memory constraints on single GPU cards, we opted for a multi-scale system with different sizes of intensity patches. The three orthogonal patches significantly outperformed the individual 2D or 3D patches, proving that they are an excellent trade-off to capture 3D information with considerably less memory than a dense 3D patch. We introduced two other input vectors ensuring the global spatial consistency of the segmentation. First, we showed that the network can learn surprisingly well the relative positions of the regions with large raw downscaled 2D patches of voxel intensities. Second, we showed that distances to centroids, despite their imprecision, provide robust inputs to efficiently capture the location of the voxel in the brain. Their robustness is due to their redundancy and their independence to translations and rotations. In our approach, global consistency was therefore enforced by the inputs of the model without having to resort to any complicated post-processing, such as conditional random fields, which are commonly used in the literature. We also observed that distances to centroids and downscaled patches contain redundant information and one may consider only one of these two sets of features and obtain almost the same performance. On the one hand downscaled patches have the advantage of using the raw averaged intensities, while distances to centroids require a preprocessing step to approximate the centroids by using two networks. On the other hand downscaled patches take a significantly larger amount of memory than the $134$ distances to centroids. Remaining errors are rather due to local imprecision of the segmentations and lie on the boundaries of the regions (cf. figure~\ref{brain:ultimate_diff}). Future work should focus on improving this local precision by, for example, sampling more training data points along these boundaries.

We obtained our best model by optimising the mean dice coefficient indirectly by considering the plain negative log-likelihood as cost function. However, we believe that future research should consider more sophisticated cost function that can take into account the high class imbalance of the problem (in the MICCAI dataset, the smallest region accounts for only $0.01\%$ of the brain volume, whereas the biggest accounts for $16.9\%$). We carried out a few experiments in which we weighted the error of each datapoint with penalty terms such as $\frac{V_{true}}{V_{pred}}$ ($V_{true}$ and $V_{pred}$ are respectively the volumes of the true and predicted regions of the datapoint) but these attempts did not lead to substantial improvements. We also tried to sample the same number of voxels per anatomical region but it did not seem to improve the performance.

In our experiments, we obtained good validation results by training huge networks, sometimes composed of tens of millions of parameters, with a relatively small amount of data (a few thousands already provided decent results). Although the trained networks overfit the training data, they still generalise fairly well to unseen MRIs. The reason is likely due to the fact that, contrary to natural images, brain MRIs are highly structured and there is relatively little variability between regions from one brain to another. This may explain why relatively few voxels are sufficient during training. Using more training atlases in order to capture more variability during training would be the ideal solution to reduce overfitting, and we expect that substantial improvements can be achieved by increasing the number of training atlases. Unfortunately, atlases are rare and expensive to obtain. Further work should consider generating artificial atlases from the existing ones by applying small transformations such as rotations, scaling, noise or other small distortions that may be plausible in real MRIs. Instead of creating whole new artificial atlases, these transformations could be included in the model itself as we did with our noise layer that corrupts distances to centroids on the fly.

{\small
\bibliographystyle{ieee}
\bibliography{biblio}
}

\end{document}